# Modelling dynamic systems transfer functions from events in computational neuromorphic imaging

Nimrod Kruger
*International Centre for Neuromorphic Systems*
*Western Sydney University*
Sydney, Australia
n.kruger@westernsydney.edu.au

Gregory Cohen
*International Centre for Neuromorphic Systems*
*Western Sydney University*
Sydney, Australia
g.cohen@westernsydney.edu.au

***Abstract*—Event Vision Sensing (EVS) report threshold crossings of log-irradiance, so a static optical system imaging a static scene produces no output at all. The classical procedure for measuring a Point Spread Function (PSF), illuminating the system with a constant point source, therefore has no event-based equivalent: the probe must carry a temporal profile, and that profile becomes part of the measurement. A growing body of Computational Neuromorphic Imaging (CNI) work already exploits this, pairing engineered or modulated optics with event sensing, but each system adopts a particular excitation together with a particular reading of the event stream without the correspondence between the two being stated. We examine that correspondence directly within a analytical framework of an Linear Shift-Invariant (LSI) optical system with a specified Modulation Transfer Function (MTF), a first-order filter EVS pixel model, and three different temporal probes: a step function, a linear ramp and an exponential ramp. By analysing the inverse of the entire chain for different event-statistic, and comparing the results to the specified MTF, we identify the context where each probe is most relevant. We consider how photon-noise and cross-array threshold mismatch effects the analytical accuracy of the probe-inverse. Results show that the widely used step probe is highly susceptible to mismatch while resilient to photon shot-noise, while a linear rise probe and exponential rise probe retain their ability to infer signal levels even with high mismatch. We discuss the potential of dynamic-PSFs as components of a full forward operator from scene to events. In this, we use this analytical description to define dynamic-PSFs around EVS, and discuss the gaps toward a unified pixel model and a scene-composition framework required for CNI.**

## I. Introduction

### A. Computational imaging with event sensors

Computational Imaging (CI) infers scene information by jointly designing an optical projection and the computation that inverts it [1]. The successful optical system design will produce a scene projection so information of interest becomes separable, ideally low-rank, under noisy measurement and with tractable compute. In this respect CI differs from signal processing, which handles temporal signals, and from image processing, which handles spatial ones, in that it must treat the full spatiotemporal problem while carrying the additional abstraction layer of the optical hardware.

EVS, or neuromorphic sensing, reports asynchronous per-pixel threshold crossings of log-irradiance [2], [3], rather than an integrated intensity value. The sampling is sparse, continuous in time and high in dynamic range, at the cost of limited sensitivity, since contrast thresholds are set relatively high to accommodate hardware mismatch [4]. This sampling mode fits neither the static image nor the isolated time-series picture, and it can be argued that neuromorphic sensing constitutes a CI problem from the outset, even behind a conventional imaging lens [5]. Challenging the narrative of CI presented above, now the on-sensor asynchronous log-irradiance threshold-crossing sampling modes introduces an additional layer of abstraction to the design, while the output sensor data is not trivially sparable in space and in time as with most CI approaches.

The emerging body of work for considering EVS-CI, collectively termed CNI, now spans multiple domains and applications. Noted, a property of EVS is that a static optical system imaging a static scene produces no events. Therefore, three distinct CNI approaches can be mentioned: Internally supplied system dynamics appear as coded apertures switched during a single measurement [6], [7] and as steered or multiplexed ray paths for light-field capture [8]. Active illumination underlies event-based light-field microscopy [9]. Information extracted from scene dynamics alone covers wavefront sensing [10] and image reconstruction posed as a linear inverse problem [11].

The promise of CNI is the ability to uniquely deliver high bandwidth and low latency scene information, not readily available by frame-based imaging systems, often with imposed design constraints. The question raised when developing such systems is “how to design an optical projection and algorithm for maximal separability of the event-statistics?”. These event-statistics must represent both the spatial and temporal behaviour of the camera, however, the literature traditionally addresses either-or.

### *B. The gap addressed*

PSF engineering has recently been extended to event cameras for 3D localisation and tracking, including task-specific mask design optimised against Cramér-Rao bounds [12]. However, this tool gets only a light touch in current literature, partly due to an incomplete picture of its role. Dynamic phase masks have also been used to build time-averaged PSFs whose attainable set is richer than that of any static mask [13]. There, the optics themselves vary in time and the variation is integrated away by frame exposure, which is the opposite limit to the case considered here. Event sensing has been combined with Fourier light-field microscopy for ultrafast volumetric imaging [9], where a 3D PSF and its spectrum are recovered from events generated by a translating bead. There the probe is uncontrolled motion, and the statistic is not chosen or inverted, so no correspondence between excitation and statistic is established.

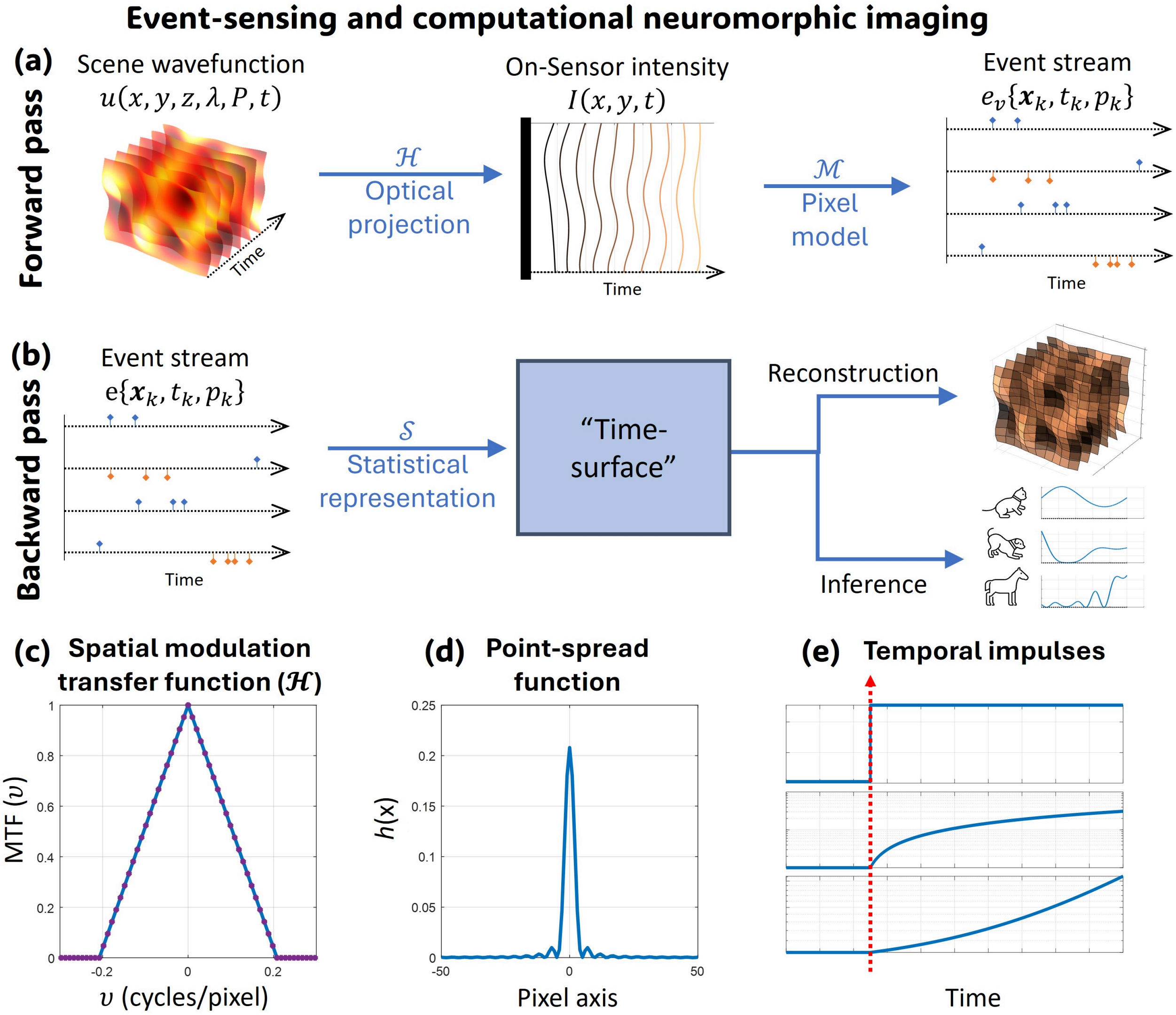

Fig. 1. Computational Neuromorphic Imaging framework, and the role of spatial-temporal probes in it. (a) A scene is projected through the optical components onto the Focal-Plane Array (FPA), and the corresponding spatial temporal signal is converted into an event-stream. This forward pass can be described by a spatial operation in the optics ($\mathcal{H}$) and a temporal point-process operator through the event-pixel model $\mathcal{M}$. (b) A backward pass must then disassociate these spatial and temporal operations by observing the statistics of the event-stream (often coined the “time-surface”), to successfully perform any inference or reconstruction task of the input scene. Useful tools to probe the relation between event-stream statistics and the spatial and temporal operations are the (c) MTF (related to the Optical Transfer Function (OTF)), (d) its corresponding PSF, and the (e) temporal impulse response. Shown here is an example of a trivial 1D rectangular pupil, giving a sinc-squared intensity PSF and a triangular MTF, and a selection of temporal impulses - a step function, a linear rise, and an exponent rise (all presented in log-intensity)

Notably, utilising a PSF needs to pair a particular temporal excitation with a particular way of reading the event stream. Some present accumulated events / event-rates as the log-intensity output equivalent, others consider inter-event intervals as intensity or log-intensity derivative, and others yet consider last-event timing when relating to a reference modulated illumination source. Event representations, sensor models and PSF engineering have each been studied in isolation [3], [14], [12]. What has not been examined is the connection between event statistics, given various temporal signals, and the spatial transfer function - and accordingly, the inverse to recover it. In the broader CNI context, as in Fig. 1(b), the event-stream statistics are reduced to a specific time-surface, while Fig. 1(a) shows the transfer from scene irradiance through an optical operator, $\mathcal{H}$, and then through a pixel model, $\mathcal{M}$ (performing a point-process realisation from FPA-irradiance). Building a time-surface from only a single statistical tool from the list above (accumulation, Inter-Spike Interval (ISI), etc.), and without a mapping to the specific type of forward operators, $\mathcal{H}$ & $\mathcal{M}$, is a gap the literature is yet to address.

The aim of this work is to discuss the potential of various event-statistics that may populate a time-surface (and thus represent the PSF of the optics and sensor), and how these correspond to various spatio-temporal probes. While such a probe would be a full spatio-temporal function, we lean into the order-wise separation of these forward operators through which the signal passes – first the optics, and then the sensor – and suggest a sparable spatial-temporal function for constructing each probe.

We specify an optical system by a given MTF (Fig. 1(c)) and derive the corresponding PSF, $\mathcal{H}$ (Fig. 1(d)). For a simple imaging optics, a point source Dirac-delta function is best suited to spatially examine pixel response, with a variety of temporal function explored. The expected event-stream output is simulated with a simple first-order filter EVS model, $\mathcal{M}$, where the only noise source is the photon shot noise. For simplicity, the analysis is performed on a 1-D spatial domain. We follow three probes, as in Fig. 1(e): a step-function, a rising linear intensity, and a rising exponential intensity - all triggered at a known time, $t_0$, and starting from a finite background intensity value, $I_{\text{bg}}$. By identifying the relation where each statistics becomes a natural interpretation of parameters of $\mathcal{H}$ and $\mathcal{M}$ - we may observe inverse functions that help relate events to either scene information, or identifying our system (retrieving the MTF, or characterising sensor parameters). The validity of this inverse is tested through the reconstruction of the specific MTF input.

This work represents an important step towards a full mapping framework from scene-components input to event statistic, and can inform joint sensor and optics characterisation, design of useful optical projections for various EVS pixel models, develop performance evaluation tools, and design efficient processing pipelines for CNI applications.

## II. A dynamic imaging model

We treat the optical system as an operator $\mathcal{H}$ mapping a scene distribution $U(\boldsymbol{r},t)$ to the irradiance on the FPA. Where the system is a LSI, this operator is a convolution with a PSF kernel $h(\boldsymbol{r},t)$,

$$I(\boldsymbol{r},t) = \mathcal{H}[U](\boldsymbol{r},t) = (h * U)(\boldsymbol{r},t), \tag{1}$$

where $*$ denotes spatial convolution and the kernel is permitted to depend on time. The analysis is restricted throughout to a spatially shift-invariant, irradiance-linear optical system. Extending it to space-variant optics would replace the convolution in equation (1) with the corresponding space-variant operator.

Because an event sensor responds to temporal change, the quantity of interest is not $I$ but its temporal derivative. Differentiating equation (1) gives

$$\frac{\partial I}{\partial t} = \frac{\partial h}{\partial t} * U + h * \frac{\partial U}{\partial t}, \tag{2}$$

so the irradiance change at any pixel receives a contribution from the optical system evolving and a contribution from the scene evolving. Both terms produce events, and the event stream alone does not distinguish them. The probe, in essence, is a representation of $U$, and when no dynamics in the optics, we are left with the second term only and must ensure our scene is dynamic to get a signal. The literature mostly considers a translating point-source when doing analytical derivations for this problem, and as will be discussed later, this imposes challenges and restricts the range of what can be derived.

The pixel converts irradiance to events in three stages. The photoreceptor is logarithmic, so the sensed quantity is

$$L(\boldsymbol{r}_i, t) = \log I(\boldsymbol{r}_i, t), \tag{3}$$

where $I(\boldsymbol{r}_i, t)$ is understood as the irradiance spatially-integrated over the active area of pixel $i$. The source-follower buffer then applies a low-pass response. Circuit-derived models generally require a second-order description, with the photoreceptor and buffer contributing separate poles [15], with one pole typically dominating [16], and intensity dependant [17], [18]. In this analysis we retain only one dominant pole,

$$\tau_{\text{pr}}\dot{V} + V = L, \tag{4}$$

with $\tau_{\text{pr}}$ representing front-end time constant. We emphasise the order of these two stages: the logarithm precedes the filter, so $V$ is a filtered log-irradiance and not the logarithm of a filtered irradiance.

The change detector compares $V$ against a stored reference. Neglecting second order pixel latency mechanisms, an event is emitted at the first instant the departure reaches the contrast threshold,

$$t_k = \inf\{t > t_{k-1} + \rho : |V(\boldsymbol{r}_i, t) - V_{\text{ref}}| \geq C\}, \tag{5}$$

with polarity $p_k$ given by the sign of the departure. Here $\inf\{\}$ denotes the infimum, and reads as the first instant at which the threshold is reached, while also covering the case where no crossing occurs with $\inf\{\varnothing\} = +\infty$. The pixel is then inactive for a refractory period $\rho$, after which the reference is updated to the current value,

$$V_{\text{ref}} \leftarrow V(\boldsymbol{r}_i, t_k + \rho). \tag{6}$$

For convenience, thresholds are taken to be symmetric, $C_{+1} = C_{-1} = C$, as we will mostly discuss monotonously rising signals. The output is the event stream $e_v = \{e_k\}$ with $e_k = (\boldsymbol{r}_k, t_k, p_k)$. We note that the reset convention in relation (6) costs one refractory period of signal per event: under a constant "drift" signal $g$ in $V$, successive events are separated by

$$\Delta t = C/g + \rho \tag{7}$$

rather than by $C/g$, and offsets every statistic reported below by a predictable amount.

The one noise source retained is photon shot noise, on the grounds that it belongs to the light rather than to any particular circuit model. Photon arrival is Poisson, so the count collected in an interval $\Delta t$ is

$$N(\boldsymbol{r}_i, t) \sim \text{Poisson}(\Phi I(\boldsymbol{r}_i, t)\Delta t), \tag{8}$$

with $\Phi$ the photoelectrons per second per unit irradiance. This noise enters before the logarithm and before the filter.

We omit, for now, event latency, higher-order front-end dynamics, and all noise sources other than photon statistics. In-depth pixel-model treatment [15], [16], [19], [18], [20] can help develop the details of how each statistics carries the optical weight - while the aim of this work is to treat analytical statistical relations between input and output, and perhaps assist in developing newer models [21]. We *will*, however, discuss aspects of threshold non-uniformity across the array [22], and how they may introduce MTF errors, as this aspect is mostly overlooked in the literature.

## III. Probes and event statistics

### A. *Specifying the system and the probe*

For an incoherent LSI system the OTF is the Fourier transform of the intensity PSF, normalised at zero frequency [23],

$$\hat{H}(\nu) = \frac{\mathcal{F}\{h\}(\nu)}{\mathcal{F}\{h\}(0)}, \tag{9}$$

and the MTF is its modulus, $M(\nu) = |\hat{H}(\nu)|$. The normalisation at zero frequency, dividing out the total flux transferred by the system, ensures the MTF describes how contrast, rather than absolute irradiance, is transferred as a function of spatial frequency.

The PSF and the MTF are therefore two representations of one quantity, and either may be measured directly (A point source for PSF and sinusoidal target for MTF). This equivalence is a result of the linearity in irradiance on the FPA, which is a property of the LSI optics. However, for EVS a direct access to each will require an additional step. We will later examine each probe's statistical projection by its ability to reproduce a linear transformation from a PSF-based measurement to a MTF. For a one-dimensional diffraction-limited system with a clear rectangular pupil the MTF (Fig. 1(c)) is the autocorrelation of that pupil,

$$M(\nu) = \max(0, 1 - |\nu| \; /\nu_c), \tag{10}$$

with $\nu_c$ the cutoff frequency, and the corresponding intensity PSF (Fig. 1(d)) follows by inverse transform as

$$h(x) = \nu_c \mathrm{sinc}^2(\nu_c x). \tag{11}$$

We adopt equations (10) and (11) for every numerical result below. This will supply a ground truth against which each statistic can be benchmarked, by recovering the original MTF from the spatial Fourier transform on a function of the event statistics.

The classical PSF measurement illuminates the system with a constant point source and records the intensity distribution on the sensor. For an event sensor the probe must also carry a temporal profile, and we write it as a spatial Dirac impulse at $\boldsymbol{r}_0$ modulated by an as-yet-unspecified profile $d(t)$,

$$D(\boldsymbol{r}, t) = \delta(\boldsymbol{r} - \boldsymbol{r}_0)\, d(t). \tag{12}$$

Acting with the optical operator, and adding a uniform ambient background $I_{\mathrm{bg}}$, gives the focal-plane irradiance

$$I(\boldsymbol{r}, t) = h(\boldsymbol{r} - \boldsymbol{r}_0)\, d(t) + I_{\mathrm{bg}}. \tag{13}$$

We refer to $I(\boldsymbol{r}, t)$ as the spatio-temporal point response under the probe $d(t)$, and to $w_i = h(\boldsymbol{r}_i - \boldsymbol{r}_0)$ as the optical weight at pixel $i$. Recovering the set $\{w_i\}$ is what is meant here by measuring the PSF from events.

The background, $I_{\mathrm{bg}}$ in equation (13) is key to ensuring the weight $w_i$ is not cancelled out during the log-differentiation at the pixel. Always $I_{\mathrm{bg}} > 0$, otherwise the weight becomes an additive constant that every difference cancels. In actual measurements, $I_{\mathrm{bg}}$ is always strictly positive as a pixel will always have at least a dark current floor, $I_{\mathrm{dark}}$, to sit on. Stated physically, an event sensor cannot see static multiplicative gain.

Now, each pixel sees a one-dimensional signal whose temporal shape is the known profile $d(t)$, and whose only free parameter is $w_i$. The pixel model of Section II converts that signal to the event stream, $\mathcal{M} : I(\boldsymbol{r}_i, \cdot) \to e_v(\boldsymbol{r}_i)$, from which we read a statistic

$$\boldsymbol{S}_i \equiv S[e_v(\boldsymbol{r}_i)] \in \mathbb{R}^m. \tag{14}$$

The statistic is computed directly from event times and polarities; counts, intervals, latencies, a response phase, or several of these together. We use event-statistic surface for a per-pixel field of counts, intervals, latencies or related quantities. This is broader than the conventional meaning of a time-surface, which encodes recency through the timestamp of the most recent event, typically with temporal decay [14], [24].

Because the pixel model acts on each pixel independently, and because the probed signal at a pixel is parametrised by the single scalar $w_i$, the composite map from weight to statistic is a **pointwise** monotone encoding function of that weight,

$$\boldsymbol{S}_i = f_d(w_i), \quad f_d : \mathbb{R} \to \mathbb{R}^m, \tag{15}$$

with the form of $f_d$ set by the probe and by the pixel parameters. Here this encoding function maps into a $m$ -dimensional space of statistical values - however, the literature typically maps into just one, while a higher $m$-dimensional mapping is genuinely underexplored. We can then write the chain from real weight, $w_i$, to estimated weight, $\hat{w}_i$, and to a recovered OTF:

$$w_i \longrightarrow \boldsymbol{S}_i \longrightarrow \hat{w}_i = f_d^{-1}(\boldsymbol{S}_i) \longrightarrow \hat{H}_d(\nu), \tag{16}$$

with the recovered OTF

$$\hat{H}_d(\nu) = \frac{\mathcal{F}\{\hat{w}\}(\nu)}{\mathcal{F}\{\hat{w}\}(0)}. \tag{17}$$

We apply the benchmark to each statistic by evaluating $|\hat{H}_d(\nu)|$ against the specified MTF of (10). The fact that $f_d$ is a function of sensor parameters means its identification requires a sensor characterisation step. By laying out several probes and their corresponding independent statistical components, the aim is to build a toolset to enable both characterisation of sensor parameters and the optic transfer function.

The order of operations described by the chain in (16) is the substantive point. The nonlinearity acts on each pixel independently and is not a convolution, so it is invertible pixel by pixel. Because $f_d$ is nonlinear,

$$\mathcal{F}\{f_d^{-1}(\boldsymbol{S})\} \neq f_d^{-1}(\mathcal{F}\{\boldsymbol{S}\}), \tag{18}$$

so a Fourier transform applied to the raw statistic yields the transform of a warped PSF, which is not the OTF and does not converge to it as noise is reduced. Inverting first and transforming second recovers the OTF, often at the cost of requiring the probe and the pixel parameters to be known. We will discuss in detail what each statistic requires in that regard.

We note an important aspect of this chain: the pointwise property of relation (15) holds because the probe supplies the entire temporal variation, so that a pixel's history is abstracted out. This allows the typically stateful behaviour of $\mathcal{M}$ to be disentangled, and directly relate the measured time-surface to the optical weight and the pixel parameters. And although $w_i$ is the elementary component from which an extended object would classically be composed, recovering $\{w_i\}$ does not by itself make the equations chain (16) a scene-reconstruction procedure for a general input. Due to the log operation of $\mathcal{M}$ the response to two sources is not the sum of the responses to each. Equations chain (16) is established here first and foremost to interrogate the optics and the sensor, and discussing its role as an inverse function to restore a scene is taken up in Section V, as a hypothesis.

### *B. The step probe*

The simplest probe is a step in irradiance,

$$d(t) = A\,u(t - t_0), \tag{19}$$

which is a standard stimulus for measuring contrast threshold and latency in event sensors [25], [2], and the closest analogue of the classical constant point source illumination used in static image pipelines. Its derivative is the impulse response, and therefore justifying its use for many EVS characterisation problems. As shown in Fig. 2(a), each pixel's log-irradiance mean makes jumps from $\log I_{\text{bg}}$ to $\log\big(I_{\text{bg}} + w_i A\big)$, and a finite burst of ON events is emitted before falling silent. The range of that step for each pixel is

$$\Delta L_i = \log\left(1 + \frac{w_i A}{I_{\text{bg}}}\right), \tag{20}$$

and every statistic below is a function of this local excitation. The natural statistic is the event count, and without accounting for the refractory period it is approximated by the range divided by the threshold,

$$N_i \approx \frac{\Delta L_i}{C}. \tag{21}$$

We may then approximate the optical weight by the inverse

$$\hat{w}_i = \frac{I_{\text{bg}}}{A}\big(e^{\Delta L_i} - 1\big) = \frac{I_{\text{bg}}}{A}\big(e^{N_i C} - 1\big). \tag{22}$$

Introducing the refractory period $\rho$ requires returning to the crossing condition of equation (5), since the reference is established only at the end of the refractory period and whatever the signal does during $\rho$ is never measured against any threshold. For the single-pole model of equation (4) the response to a step is $V(t) = \Delta L_i\big(1 - e^{-t/\tau_{\text{pr}}}\big)$, and writing $x_k = e^{-t_k/\tau_{\text{pr}}}$ turns the condition $V(t_{k+1}) - V(t_k + \rho) = C$ into the linear recursion

$$x_{k+1} = e^{-\rho/\tau_{\text{pr}}} x_k - C/\Delta L_i, \tag{23}$$

with $x_1 = 1 - C/\Delta L_i$, since no refractory window precedes the first event. The burst ends when $x_k$ reaches zero, at which point the source-follower voltage has settled and no further crossing occurs. Solving the recursion about its fixed point and taking that limit gives the expected event count,

$$N_i \approx \frac{\tau_{\text{pr}}}{\rho} \log\left(1 + \frac{\Delta L_i\big(1 - e^{-\rho/\tau_{\text{pr}}}\big)}{C}\right). \tag{24}$$

Being a monotone function, we may invert to estimate the optical weight similar to equation (22), providing us with

$$\begin{aligned} \hat{w}_i &= \frac{I_{\text{bg}}}{A}\big(e^{\Delta \hat{L}_i} - 1\big), \\ \Delta \hat{L}_i &\approx \frac{C}{1 - e^{-\rho/\tau_{\text{pr}}}}\big(e^{N_i \rho/\tau_{\text{pr}}} - 1\big). \end{aligned} \tag{25}$$

Two cascaded exponentials appear here, with the outer following the logarithmic photoreceptor, and the inner due to a refractory-limited burst for the single-pole pixel model.

For pixel $i$ we write the event times following onset as $t_{i,1} < t_{i,2} < ... < t_{i,N_i}$, all of ON polarity, with intervals $\Delta t_{i,k} = t_{i,k+1} - t_{i,k}$. Three components of $\boldsymbol{S}_i$ are considered for this probe:

- the **event count** $N_i$, of (24);
- the **onset latency** $\ell_i = t_{i,1} - t_0$, the delay of the first event from the probe onset;
- and the **first interval** $\Delta t_{i,1}$.

The aim is to show how a useful set is arrived at and what each member on this list contributes, first to probe-based characterisation and later to event-based processing pipelines. With more complex pixel models, additional statistics may be considered. Not every statistic here is defined at every pixel: the interval requires $N_i \geq 2$ and the latency $N_i \geq 1$, while a count of zero is itself a measurement.

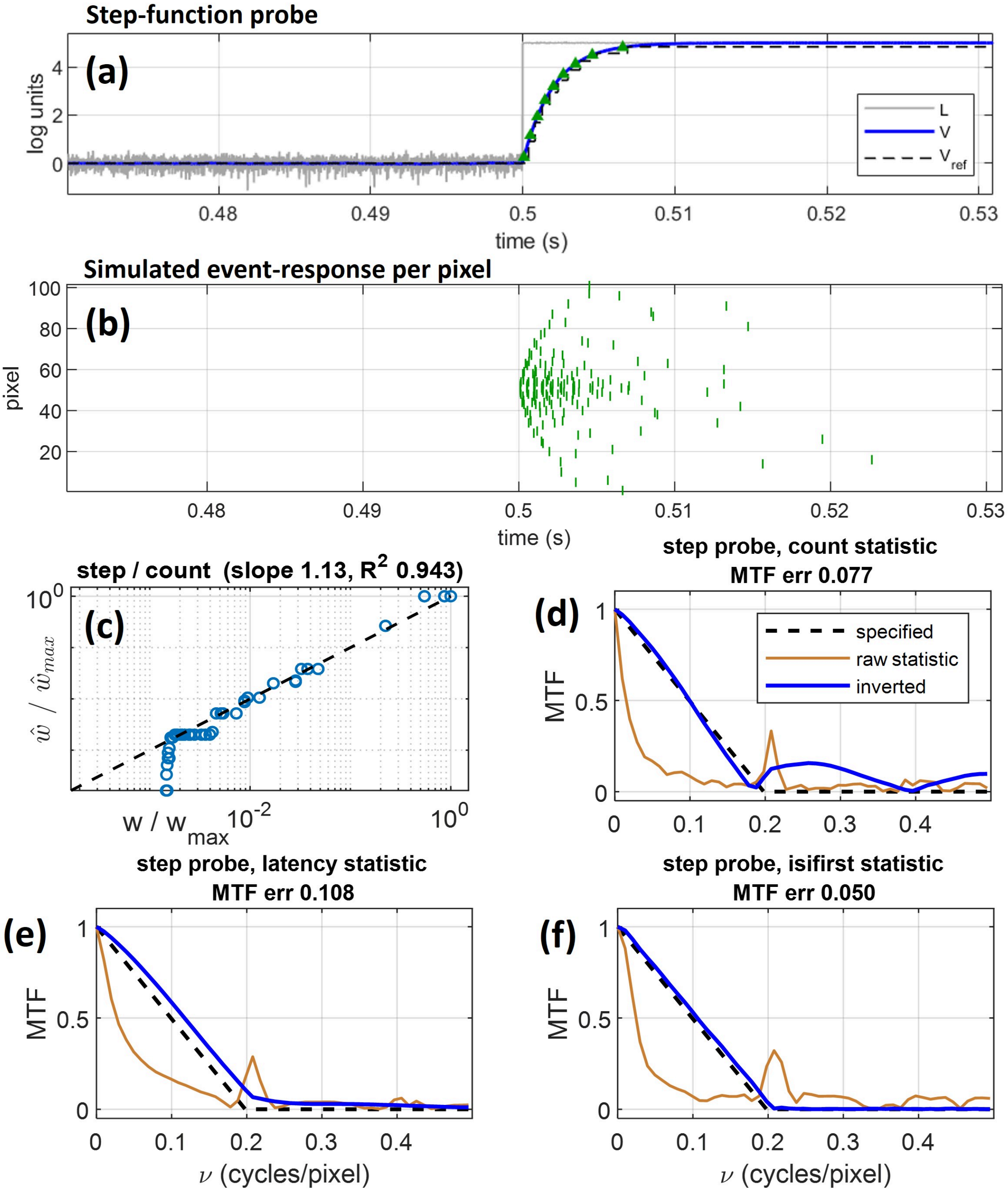

Fig. 2. A step probe. (a) The simulated log-irradiance of a pixel (gray line), its source-follower voltage (blue), and the emitted events (green triangles) with their corresponding reference voltage (dashed line) as responding to a step function increase of intensity from $I_{\text{bg}} = 2.5 \cdot 10^6$ photons/sec to $I = 1.8 \cdot 10^9$ photons/sec, and a a contrast threshold of $C = 0.25$ log units. (b) For the full 1-D sensor response to the step, with a spatial intensity distribution profile as in Fig. 1(d), where events per pixel are presented as green marks - with a maximal event count of 9 for the highest intensity step, and an overall count of 173 events over the 101 simulated pixels. (c) The pixel-intensity weight estimation (normalised) from equation (25), compared to the ground truth pixel weights, and a dashed line to rks expected for ideal reconstruction. (d) The estimated MTF from the inverted event-count from equation (17) (blue line), compared to the non-inverted attempt to measure MTF from counts (brown line), and the specified ground-truth MTF (dashed line). Similarly, the non-inverted and inverted estimation of MTF are presented for (e) the latency from step onset, and (f) first ISI. Corresponding RMS error for the estimation across the entire band is 7.7%, 10.8%, and 5% for the count-based, first event latency, and first ISI respectively.

All three recover the same underlying log range $\Delta L_i$ of equation (20), and also share the outer inverse of equation (25). They differ in how that range is read from the event stream,

$$\Delta \hat{L}_i = \begin{cases} \dfrac{C}{1 - e^{-\rho/\tau_{\text{pr}}}} \left( e^{N_i \rho/\tau_{\text{pr}}} - 1 \right) & \text{(count)} \\ \dfrac{C}{1 - e^{-\ell_i/\tau_{\text{pr}}}} & \text{(latency)} \\ C\left(1 + \dfrac{1}{e^{-\rho/\tau_{\text{pr}}} - e^{-\Delta t_{i,1}/\tau_{\text{pr}}}}\right) & \text{(first interval).} \end{cases} \tag{26}$$

The three differ in what must be known and in how they degrade. The count and the first interval each require the three sensor parameters $C$, $\rho$ and $\tau_{\text{pr}}$, as does the first interval measure. The latency fitting, however, requires only $C$ and $\tau_{\text{pr}}$, but in exchange requires the probe onset $t_0$ to be known exactly. Whether that is a saving depends on the experiment: in a characterisation rig the onset is a trigger and is free, whereas an optical system driven by an uncontrolled excitation supplies no such reference.

Observing a sample simulation output for a single pixel in Fig. 2(a), the log-irradiance transitions from $I_{\text{bg}} = 2.5 \cdot 10^6$ photons/sec to $I = 1.8 \cdot 10^9$ at $t_0 = 0.5s$ as shown by the grey line. The blue line follows the simulated source-follower voltage, smoothed by the single-pole filter, and responding to the step with a delay according to $\tau_{\text{pr}}$. The emitted events are then plotted as green triangles whenever $V$ crosses a contrast threshold value, set here to be 0.25, and a new static reference voltage is set (dashed line) after an additional $\rho = 300\mu s$. For an array of 101 pixels exposed to this step response through simulated optical operator $h(x)$ as in equation (11), we get a raster plot of events across time and pixel-space in Fig. 2(b).

With the event-count statistic, and using the inverse of equation (25), we recover the illumination weight per pixel, $w_i$. Normalising this value to the maximal $w_i$ and plotting against the ground truth, as shown in Fig. 2(c), we note a good fit generally for higher values and reduced performance for lower weights, fitting the floored-value nature of an event count for low signals. Nonetheless, reconstructing the MTF using equations (16) - (17), as shown in the blue line in Fig. 2(d), yields an estimate of the original MTF (dashed line) down to a 7.7% Root Mean Square (RMS) error, and considerably better than the attempt to naively infer the MTF from counts alone. Higher frequency artefacts are expected, as evident from the sidelobes of the reconstructed MTF. For a MTF reconstruction using the onset latency alone we get an RMS error of 10.8% - resulting from a small variations in latency (due to photon shot-noise) being amplified through the $\Delta \hat{L}_i$ relation, but with evidently reduced high-frequency sidelobes in relation to the count (Fig. 2(e)). Highest performance is achieved with the first ISI, reaching a 5% ground truth fitting to MTF with the same data, and no sidelobes. however, all these reconstructions are highly unstable when variations in probe parameters and noise is introduced, as will be discussed in Section IV, making this a finicky tool at best.

### C. *The linear ramp*

A linear rise,

$$d(t) = \alpha\,(t - t_0), \tag{27}$$

produces a log-irradiance that rises monotonically but with an ever-decreasing slope. The major difference from the step response is that a pixel continues to emit events throughout the record, which is both a blessing and a curse: realising this probe in the laboratory requires a source with adequate range, but we will see that $w_i$ is better estimated for a comparable event count to that of the step-probe.

Writing $s = t - t_0$ and collecting the pixel-dependent terms into a characteristic time

$$T_i = \frac{I_{\text{bg}}}{w_i \alpha}, \tag{28}$$

the log-irradiance at pixel $i$ becomes

$$L_i(s) = \log I_{\text{bg}} + \log(1 + s/T_i), \tag{29}$$

a function of $s/T_i$ alone. In contrast to the step probe in equation (20), where the optical weight sets the *amplitude* of the excursion, under a linearly rising ramp it sets its *time scale*. The input signal change itself is unbounded, so there is no equivalent of $\Delta L_i$ to recover, and every statistic below estimates $T_i$ instead.

Applying the crossing condition of equation (5) to equation (29), with the reference taken at the end of the refractory window, gives

$$\frac{T_i + s_{k+1}}{T_i + s_k + \rho} = e^C. \tag{30}$$

To extract how this evolves over the event series per pixel we define the variable $u_k = T_i + s_k$ as the linear recursion $u_{k+1} = e^C(u_k + \rho)$. Before onset the irradiance is constant, so the reference has settled at $\log I_{\text{bg}}$ and the first event carries no preceding refractory window. The initial condition is therefore $u_1 = T_i e^C$, which is the statement that $s_1 = T_i\big(e^C - 1\big) > 0$. Solving the recursion about its fixed point gives $u_k = \tilde{T}_i e^{kC} - \rho e^C/\big(e^C - 1\big)$ and hence the ISIs

$$\Delta t_{i,k} = \tilde{T}_i\, e^{kC}\big(e^C - 1\big), \quad \tilde{T}_i = T_i + \frac{\rho}{e^C - 1}. \tag{31}$$

It is worth being explicit about where the pixel model enters (31). The crossing condition is applied to the log-irradiance of (29) rather than to the source-follower voltage, under the assumption that the probe is slowly varying. This means as $\tau_{\text{pr}} \ll T_i + s$, the single pole filter of equation (4) acts as a pure delay, $V(s) \approx L\big(s - \tau_{\text{pr}}\big)$. What follows is that while the source-follower pole-related delay is common to every crossing at every pixel (assuming array uniformity), it cancels between consecutive events, as long as the slow slope condition applies. The practical outcome is that whenever it is not (steep slope signals), we may just wait, and consider events at regions where the slope flattens enough.

Inverting (31) at the first interval gives

$$\hat{w}_i = \frac{I_{\text{bg}}\, e^C\big(e^C - 1\big)}{\alpha\big(\Delta t_{i,1} - \rho\, e^C\big)}. \tag{32}$$

No exponential of a measured quantity appears. Where the step required the amplitude of (20) to be brought back through the logarithm, a time scale is read from event times directly, and the inverse is affine in the measurement followed by a reciprocal.

As in Section III.B we take three components of $\boldsymbol{S}_i$, now the first interval $\Delta t_{i,1}$, the onset latency $\ell_i = t_{i,1} - t_0$, and the mean inverse interval, or event rate, $R_i = (N_i - 1)^{-1} \sum_k 1/\Delta t_{i,k}$. Their inverses follow from (31),

$$\hat{w}_i = \frac{I_{\text{bg}}}{\alpha} \cdot \begin{cases} \dfrac{e^C\big(e^C - 1\big)}{\Delta t_{i,1} - \rho e^C} & \text{(first interval)} \\ \dfrac{e^C - 1}{\ell_i - \tau_{\text{pr}}} & \text{(latency)} \\ \big[\Gamma(N_i)/R_i - \rho/\big(e^C - 1\big)\big]^{-1} & \text{(rate)} \end{cases} \tag{33}$$

$$\Gamma(N) \equiv \big(1 - e^{-(N-1)C}\big)/\Big((N - 1)\big(e^C - 1\big)^2\Big).$$

Only one number in each of these inverses has to be known. The recovered OTF of equation (17) is normalised at zero frequency, so any factor shared by every pixel divides out before the transform is taken. In the inverse set (33) the perfector $I_{\text{bg}}/\alpha$ and every numerator are shared in this way. Recovering the transfer function under a ramp therefore requires a single additive time offset: $\rho e^C$ for the first interval, where the front-end delay has cancelled and only the reset convention remains; $\tau_{\text{pr}}$ for the latency, where the delay survives and the reset convention has not yet acted; and $\rho/\big(e^C - 1\big)$ for the rate. This is in contrast to the step probe, that requires a pixel model to be recovered as the exponential operation doesn't scale across pixels, and is not normalised when examining the MTF. Considering, however, more realistic pixel models with multiple-pole response, using steep linear slopes, or starting from low illumination signal, does break some assumptions mentioned earlier. Therefore, examining the full interval histogram and change over time may prove more useful for system identification problems.

We also note that the inverse needed for the linear probe may often be small. For the first interval, for instance, the offset is $\rho e^C$. At the parameters used here it is no more than 4% of $\Delta t_{i,1}$ at the brightest pixel, and less at every other, since dimmer pixels have longer intervals. The reciprocal first interval is therefore already close to proportional to $w_i$, and the inverse removes a small bias rather than a warp. That bias grows with the ramp rate: a faster ramp shortens the measurement time but makes knowledge of $\rho$ more important.

For the latency the offset is $\tau_{\text{pr}}$, roughly 6% of $\ell_i$ at the brightest pixel for representative values chosen in our simulation. The latency thus depends on its inverse more than the interval does. It returns something for that dependence:

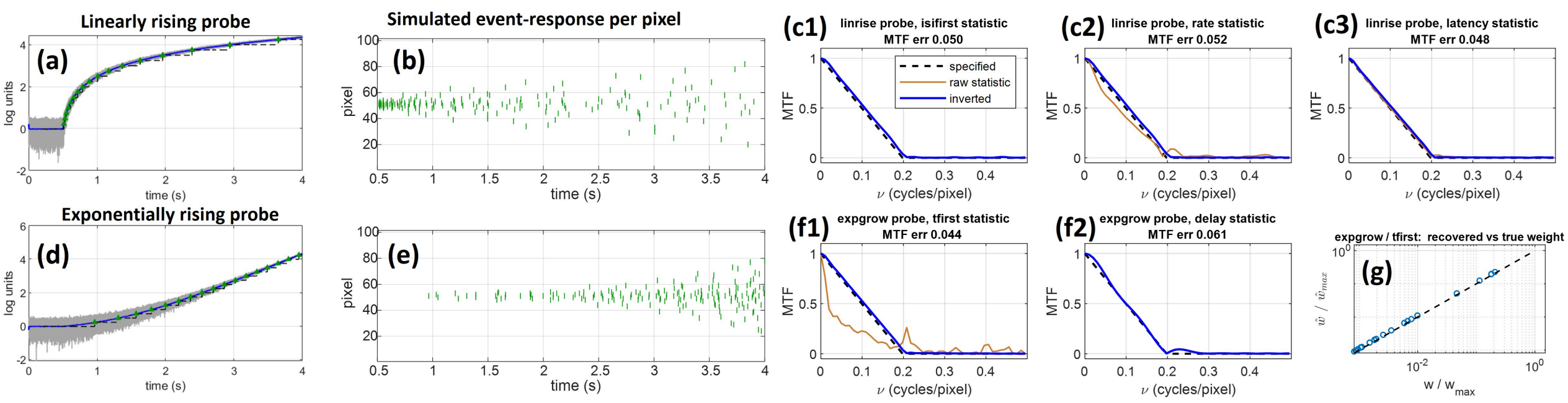

Fig. 3. Monotonically rising probes. (a) For a liear rising probe in log-irradiance values, we see continuously emitted events with ever increasing intervals. (b) For a full 1-D sensor simulated response to the input through $h(x)$ the events per pixel are presented as green marks in relation to time. Measurement time and slope value was tuned to produce the overall same count of events as in Fig. 2 - with a maximal event count of 17 for the highest intensity step, and an overall count of 174 events over the 101 simulated pixels. Estimation of the specified MTF is shown using (c1) the first interval, (c2) mean rate, (c3) and first event latency, with error RMS roughly 5% for all those methods. For each, we present the inverse-function based estimation (blue line), and direct-from-event estimation (teal-line), compared to the specified ground truth (where teal-line not present, it is underneath the blue line, indicating no inverse is needed in practice). (d) Similarly, an exponentially rising input signal is analysed, (e) also tuned to have a similar event count across the array (17 events at peak pixel, and 172 event in total), and an MTF reconstruction error of (f1) 4.4% for first interval inverse, and (f2) 6.1% for first event delay. (g) Presenting the optical weight estimated from the first interval shows ideal match to the expected value for each pixel, indicating the MTF error is only attributed to discretisation error (sub-threshold pixels).

plotting $\ell_i$ against $1/\hat{w}_i$ gives a straight line whose intercept is $\tau_{\text{pr}}$, so a ramp measurement can also report the front-end time constant.

The rate statistic differs in its role. Every interval is proportional to the same $\tilde{T}_i$, so no interval beyond the first carries new information about $w_i$, and averaging over them acts to only reduce the variance of the estimate. Averaging does introduce one dependence: the reciprocal intervals form a geometric series that saturates after a few terms, while the divisor $N_i - 1$ keeps growing with the observation window. The factor $\Gamma(N_i)$ in equation (33) corrects for this. The correction is mild, since $N_i$ varies only logarithmically with $w_i$ while $\tilde{T}_i$ varies over decades, so the raw rate already tracks the weight closely. What might be uncovered through tracking this rate, are higher-order pixel model parameters, and specifically those illusive intensity dependant ones.

Compared with the step probe of Section III.B, the ramp recovers the MTF more accurately from every statistic considered, and does so without the quantisation that limits an integer-valued count. The reason is visible in equation (33): each estimate is a continuous function of a measured time, so its resolution is set by the temporal precision of the sensor rather than by the contrast threshold. The cost is that the probe must be constructed so the ramp is sustained over the measurement period.

### D. *The exponential ramp*

An exponentially rising irradiance,

$$d(t) = A\left(e^{(t-t_0)/\tau_\alpha} - 1\right), \tag{34}$$

is the probe matched to a logarithmic photoreceptor. Writing $s = t - t_0$ as before, the irradiance at pixel $i$ is $I_i(s) = I_{\text{bg}} + w_i A\left(e^{s/\tau_\alpha} - 1\right)$, and the sensed quantity is

$$L_i(s) = \log\left[\left(I_{\text{bg}} - w_i A\right) + w_i A\, e^{s/\tau_\alpha}\right]. \tag{35}$$

Once the second term dominates, this tends to

$$L_i(s) \to \log(w_i A) + s/\tau_\alpha. \tag{36}$$

The consequence is the defining property of this probe. The log-irradiance approaches a straight line whose **slope is $1/\tau_\alpha$ at every pixel**, and the optical weight only applies an additive offset to this slope.

Applying the crossing condition of (5) to (36) gives

$$(s_{k+1} - s_k - \rho)/\tau_\alpha = C \quad \Rightarrow \quad \Delta t_k = C\tau_\alpha + \rho, \tag{37}$$

for every pixel at every event index, past the asymptotic first few events close to $I_{\text{bg}}$. The weight, however, has not been lost - it has moved into the timing of the burst. The first crossing occurs when $L_i$ has risen by $C$, and solving equation (35) at that point, without any asymptotic approximation, gives the onset latency

$$\ell_i = \tau_\alpha \log\left(1 + \frac{I_{\text{bg}}\big(e^C - 1\big)}{w_i A}\right), \tag{38}$$

which inverts to

$$\hat{w}_i = \frac{I_{\text{bg}}\big(e^C - 1\big)}{A\big(e^{\ell_i/\tau_\alpha} - 1\big)}. \tag{39}$$

As under the linear ramp, the front-end pole adds a delay $\tau_{\text{pr}}$ to $\ell_i$ once the signal is slowly varying, though it scale against the probe exponential time constant. For relatively small $\tau_{\text{pr}}/\tau_\alpha$, this delay can be made negligible. The inverse of the measured latency in equation (39) contains the factor $I_{\text{bg}}\big(e^C - 1\big)/A$, which is shared by every pixel and divides out under the zero-frequency normalisation of equation (17). Recovering the transfer function under this probe therefore requires no sensor parameter at all, to within the small $\tau_{\text{pr}}/\tau_\alpha$ assumption noted above.

The defining probe property in equation (37) can be exploited in various imaginative ways. Because it holds at every pixel regardless of illumination, the interval reports $C\tau_\alpha + \rho$ across an array whose pixels differ by orders of magnitude in the light they receive. Using this probe, for instance, with unstructured illumination (full frame - and without managing illumination uniformity) will provide a measure of $C\tau_\alpha + \rho$, and repeating the measurement with another $\tau_\alpha$, for instance, may retrieve pixel-level contrast threshold and refractory period variation across the array. Threshold, photoreceptor bandwidth and refractory period are separately bias-controlled quantities that jointly set the event rate [26], so a probe that separates them without requiring uniform illumination is of practical value.

In essence, a simple EVS model $\mathcal{M}$ under an exponential ramp divides into a part that is stationary in time, carrying only sensor parameters (equation (37)), and a phase, that carries only the optical weight (equation (38)). This combination is useful beyond this statement. For one, equation (37) can be considered a null test - and as we predict an interval independent of $w_i$, any measured dependence is a departure from the model. Following the evolution of $\Delta t_{i,k}$ will uncover instability or intensity dependant front-end bandwidth, contrast thresholds, or refractory periods. Secondly, equation (38) has a knee at $w_i A = I_{\text{bg}}\big(e^C - 1\big)$, and this can be used to measure $I_{\text{bg}}$ - especially useful when no background illumination is present to uncover dark current values. A third avenue for this tool is considered as we return to equation (2), where we asserted that our optical transfer function, $h(x)$, is static. This assumption enabled the derivation under the evolution of $U(x)$ alone, but limited the usefulness of the probe to derive response to a **dynamic-OTF** - where $\partial h/\partial t \neq 0$. With such clear functional separation of components, analysis of dynamic-OTF may be fully explored.

Practically, the exponential probe might be the least convenient to realise for characterisation purposes (potentially a current-driven LED with a logarithmic driver, or a fine geometric staircase with high bit-depth digital controller), but the above potential benefits argue for this probe's case.

## IV. Noise and non-uniformity

The analytics above considered the probes under a single noise condition, the photon rate Poisson limit, which is mostly "muted" as the value is set high enough. In this section we vary two noise sources independently and consider how they effect the relations shown above.

The first is photon shot noise, which belongs to the light itself rather than any other noise introduced by the system. Its effect on the sensed quantity is a fluctuation $\sigma_V \sim 1/\sqrt{2\Phi I_{\text{bg}}\tau_{\text{pr}}}$ on the source-follower voltage, and we report it as $\sigma_V/C$, the fluctuation in units of the contrast threshold. The second is purely sensor-level component: the contrast-threshold mismatch. As a pixel specific threshold $C_i$ is a variation around a nominal $C$ value (emulated here with normal distribution with standard deviation $\sigma_C$), the inverses of Section III.B to Section III.D are calculated for the nominal value alone. This type of "fixed-pattern noise" [22] does not average away over repeated measurements, and we report it as $\sigma_C/C$. For each condition the weight is recovered separately in every noise realisation and the MTF error computed from each, so we plot the expected error of a single measurement rather than the error for an average of many repeated reconstruction experiments.

### A. *Photon shot noise*

All statistics show a trend of degrading with $\sigma_V/C$ and fail between $\sigma_V/C \approx 1$ and 2, which is the point at which a fluctuation is comparable to the threshold itself and crossings cease to be signal driven. Two features of the ordering are worth noting.

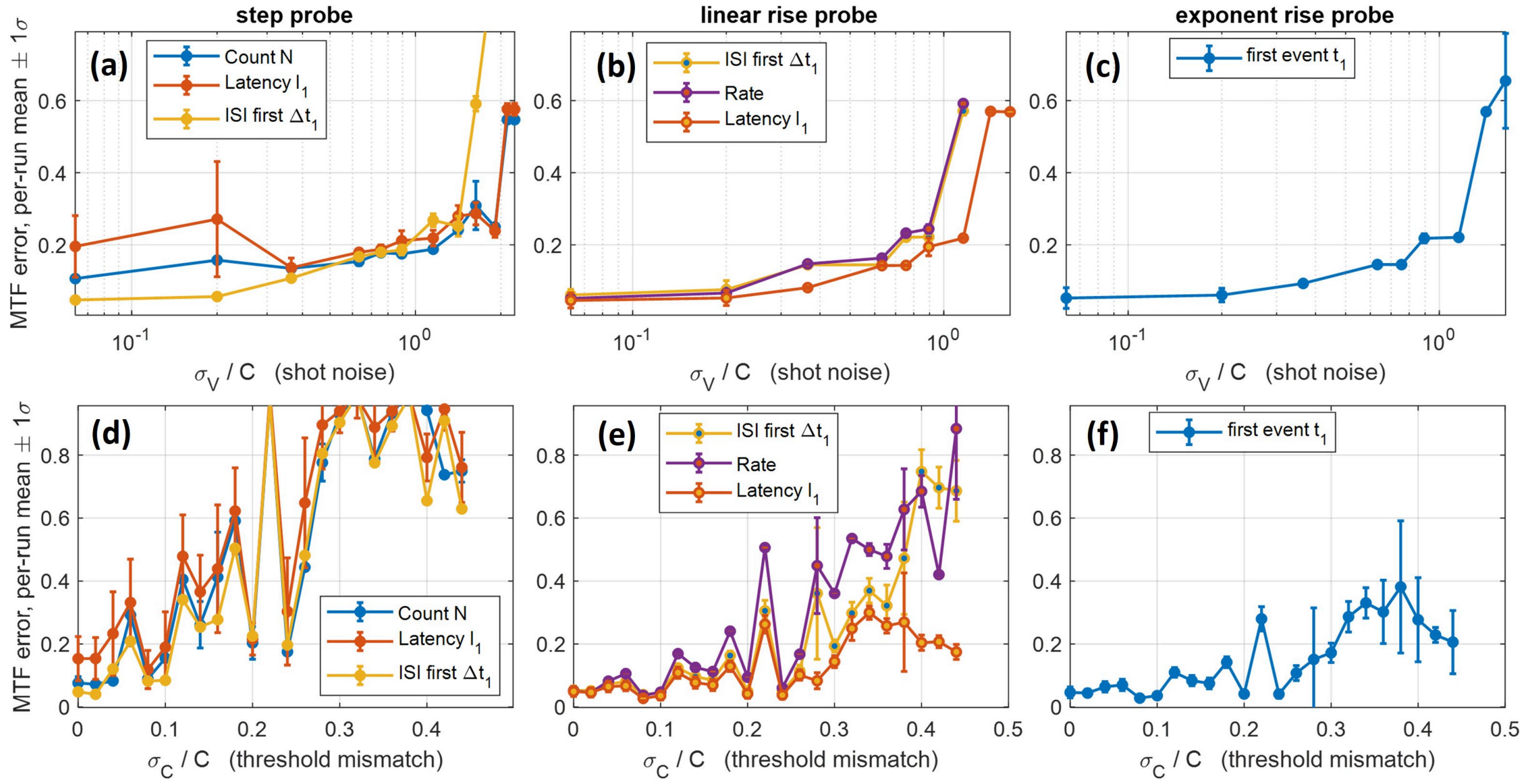

Fig. 4. Noise mode analysis for probes and their statistics. We consider the evolution of MTF estimation error with increased Poisson shot-noise for the (a) step function probe, (b) linear rise probe, and (c) exponential rise probe. The x-axis presents the standard deviation of the photon noise (estimated as the square root of the inverse photon flux), and normalised to the contrast ratio. For each we considered how the different statistical inverse chain performs, as detailed in Section III: event count $N_i$ (blue line), first latency value $l_{i,1}$ (red line), and first ISI $\Delta t_{i,1}$ (yellow line) for step probe; first ISI $\Delta t_{i,1}$ (yellow line), Rate $R_i$ (purpel line), and first latency value $l_{i,1}$ (red line) for linear rise probe; first event time $t_{i,1}$ for the exponential rise probe. Generally we see how the probes fail as the noise becomes equivalent to the contrast ratio ($\sigma_C/C \sim 1$). Similarly, we consider how cross-pixel contrast threshold mismatch effects the MTF error, with (d) a step function probe analysis failing with any variation above 5%, while (e) a linear rise probe maintaining sub 20% accuracy even with 15% mismatch values, and the (f) exponential rise performing well even with around 20% normal mismatch.

The under low noise, we see the ranking as shown in Fig. 2, where the first interval performs best, and then an event count second (bounded by quantisation), and the latency performing the lowest (based on only a single event). Then, when the relative noise reaches $\sigma_V/C \approx 0.4$ we see all three statistics reaching the same level of performance up until $\sigma_V/C \approx 1$, where now the first interval fails first.

For the ramp-based probes, the trend is monotone for all methods. We see the latency value for the linear slope performs best across all noise levels within this probe - and the first event time for the exponential rise probe following the same trend of this best performing statistic. These only fail after $\sigma_V/C \approx 1$. The other metrics for the linear rise, $\Delta t_1$ and rate, fail somewhat earlier but, in opposed to the latency metric, do not need the reference “onset time” $t_0$ to be known. Having no such “independent” test for the exponent rise, we tested the delay value from a reference first-to-spike-pixel as the “onset time”, and see a similar good performance in low noise levels, with a similar earlier failing point (data not shown).

### B. *Contrast-threshold mismatch*

This is a system-level noise and chosen for analysis with $C$ being a central parameter of any inverse function above. With this we see the step probe deteriorates very quickly, for any mismatch above with more than 5% of nominal values quickly entering the a specified MTF reconstruction error of more than 20% across all statistical analysis modes.

The reason is visible from the inverse function themselves. Under a step, the measured count carries the true per-pixel threshold, so substituting it into an inverse that assumes the nominal value gives $\Delta\hat{L}_i = \Delta L_i\, C/C_i$, and after the outer exponential

$$\hat{w}_i/w_i \approx \exp(-\Delta L_i\, \delta C_i/C). \tag{40}$$

The error is exponential in the mismatch and amplified by the log range of the probe. Alternatively, under both ramps the threshold appears in the inverse only through a factor shared by every pixel, which the zero-frequency normalisation of equation (17) removes, and the residual sensitivity, $\kappa \equiv \mathrm{d}\log(\hat{w}_i/w_i)/\,\mathrm{d}(\delta C_i/C)$, is

$$\kappa = \begin{cases} C\Big(1 + (1 - e^{-C})^{-1}\Big) & \text{first interval} \\ Ce^{C}/(e^{C} - 1) & \text{latency} \end{cases} \tag{41}$$

which is of order unity and, unlike equation (40), and independent of the probe's dynamic range.

A second consequence of equation (40) is that the per-pixel error is lognormal in the mismatch, so a step-probe measurement will not fail gradually with mismatch distribution. A particular realisation of the mismatch pattern, for instance with a scattered sensitive pixel and/or dead pixels, will prove detrimental for MTF retrieval from a step probe, and such a sensor should be evaluated through ramp-based probes.

### C. *Reading the two together*

The above shows how each probe varies with noise modes, with step-probe performing somewhat better under high photon noise, and ramps resilient to sensor mismatch, but only while light is sufficient. This bears directly on how a measured transfer function should be reported. A MTF recovered from an event stream is not a property of the optics alone; it is a property of the optics, the sensor and the probe jointly, and two measurements of the same lens through the same array under different probes will carry different sensitivity to the same non-uniformity. Recent calls for shared calibration and metadata conventions in event-based vision [21] are, on this reading, partly a request for the excitation to be reported alongside the result.

These results are indicative and are not a noise analysis. Part of the degradation seen at high noise is the loss of support (less participating pixels) rather than the corruption of the measurement. The error metric is bounded above, so it does not distinguish degrees of failure once a reconstruction has failed. What we take from Fig. 4 is the ordering of the probes and its explanation in (40) and (41), and an call to be explicit in defining noise modes, excitation methods, and pixel model whenever benchmarking EVS performance.

## V. Discussion

### A. *What is a dynamic-PSF*

The term "dynamic-PSF" might be attributed to various parts of a EVS-based system, and we start this section by unmasking this definition. As in equation (2), the full change signal on the FPA will be built from both scene dynamics and any dynamics introduced by the optical operator. Separating the two origins by definition is not an option, for three reasons. The first is the frame of reference choice, where, for instance, a translating camera is scene motion in the camera frame and optics motion in the world frame, while the event stream is identical in both. Any definition, then, must nominate a frame - and this choice is arbitrary or stems from design considerations, and not from the core definition of what a PSF is. Secondly, adding active control – when the OTF is also a function of the signal – as in adaptive optics, autofocus or closed-loop steering, the two terms of (2) are not merely simultaneous but correlated. The third is evident from (2) itself: the event stream is a functional of the sum of the two terms, and no operation applied to that stream alone splits it.

From this it follows that a dynamic-PSF is defined whenever $\partial I/\partial t \neq 0$ at fixed $\boldsymbol{r}$, irrespective of which term of equation (2) supplied the derivative. The price of this choice is that the PSF ceases to be a property of the optics alone, so every use of it must be accompanied by the excitation that produced it. When working with frame-based sensors, this aspect didn't often pose an issue, as the senor integrated over a time-frame, and any changes in the optics where instantaneous and time-averaged [13]. However, for EVS there is no such luxury, and a dynamic-PSF should be reported together with the origin of its dynamics, scene-based, optics-based, or both.

### B. *From a probe to a scene*

The reprocussions of the fused definition above is most sever when considering the classical signals-and-systems approach used to describe a LSI system only by its response to specific probes, and then predict its response to any input. In CI, for instanse, a PSF is a powerful tool used to represent the optics' response to a "component" of a scene, and this is often used to computationally predict the response of a full multi-component scene and, moreover, to later solve the inverse problem.

In this work, the chain of equation (16) holds because the probe supplies the entire temporal variation, and because every pixel of the array shares a single onset $t_0$. Those two conditions make the encoding function of equation (15) pointwise in the optical weight and any signal with a shared onset time, such as can be manufactured by a characterisation rig, satisfies both conditions. A dynamic scene, however, does not satisfy these conditions. Components at different positions carry different temporal profiles and arrive at different times, so a pixel sees a superposition

$$L_i(t) = \log\left( I_{\text{bg}} + \sum_j w_{ij}\, d_j\big(t - t_j\big) \right), \tag{42}$$

with the index $j$ running over scene components. Two properties of Section III are lost at once. The excitation at a pixel is no longer parametrised by the single scalar $w_i$, so equation (15) is a functional of the superposition rather than a function of a weight; and because the photoreceptor is logarithmic, the response to two components is not the sum of the responses to each. Recovering $\{w_i\}$ under a single probe is accordingly not a scene reconstruction procedure, and none of the inverses of Section III should be read as such.

What is missing is a composition rule. A decomposition tool requires, first, an operation on the statistic space under which the statistic of a superposition is built from the statistics of its components. This to say, we find a probe and statistic pair for which $f_d$ becomes an eigenfunction of a given event statistic, and then relating scene composition to statistic composition, rather than merely an invertible map on a single weight. It requires, second, that onset diversity enter as a variable of the framework rather than as the fixed parameter it is here. None of these are supplied by the probes listed in this work, and the progress needed to supply them is analytical rather than experimental (for the simplified pixel model at first). Third, it requires that the statistic set carry both *what* and *when*, and in a mode that is fundomentally separable. We expect that accumulative statistics that are stationary in time report only that a component is present, while phase-like statistics place it on the time axis, and a spatio-temporal decomposition needs to observe both channels at once.

Such behaviour is seen, to a limited extent, in the statistics of the exponential rise probe, and might hint at similarly derived probes to deliver this scene composition. There, a simple pixel model divides into a part that is stationary in time and carries only sensor parameters, equation (37), and a phase that carries only the optical weight, equation (38). A separation of that kind is what a decomposition needs, an amplitude-like channel that identifies a component and a phase-like channel that places it in time, obtained in a domain where the logarithmic nonlinearity has already been absorbed.

This work uses the bare probe to interrogate the optics and the sensor with no full spatio-temporal scene present. The second phase to follow on must consider a scene as a superposition of probe components with the event statistics deciding whether that decomposition is recoverable under noise. A probe chosen as an eigenfunction of a given event statistic is our conjecture for a full CNI framework.

### C. *The moving point source is a time-limited step*

We address here a widely used translating point source - where an illumination source is moved across the array, and presents each pixel with a rise and a fall of its local irradiance. Per pixel the excitation is a step up followed by a step down, with a dwell time set by the transit speed and the width of the PSF. Everything derived in Section Section III.B therefore applies, subject to two constraints that the step probe does not carry. The excursion is time-limited, so the burst may be truncated before $\Delta L_i$ of equation (20) is reached; and adjacent pixels see the same excitation at different times, so the array is not probed simultaneously and the single-onset condition used above does not hold.

A large fraction of event-camera characterisation uses a moving dot, a rotating disc or a translating target. The connection between this work to those may curtail any overreaching and generalised claims on EVS performance by a limited-step-probe test (to better of worse), while also stating where this connection work and where not. When dwell time exceeds the burst duration, the count statistic reaches the same $\Delta L_i$ as a clean step and the inverse of equation (25) applies unchanged. Where the transit is fast relative to $\tau_{\text{pr}}$ and the burst is truncated, the count saturates below $\Delta L_i / C$ and the inverse under-reports the weight. The dimmer the pixel, the earlier this bites, so the truncation is weight-dependent and does not divide out under the zero-frequency normalisation of equation (17). The recovered PSF is narrowed at its tails, and the recovered MTF correspondingly overstates transfer at high spatial frequency.

Here is an example of where such details matter. Event-based Fourier light-field microscopy calibrates a 3D PSF from an event stream generated by a translating bead, takes its spectrum as an MTF, and reports agreement with frame-based reference alongside an observed axial narrowing [9]. At the transit speed used there, the dwell-limited effect above is negligible - but two effects that are not remain: pixels below $w_{\min} = \big(I_{\text{bg}}/A\big)\big(e^C - 1\big)$ report nothing, and a statistic taken as a proxy for irradiance without the inverse of equation (18) carries the log warp. The first narrows the recovered support, the second broadens the apparent profile, and their balance depends on where the threshold sits within the profile, making the achieved correlation difficult to extrapolate to other parameter domains (faster objects, different illumination regimes, etc.).

### D. Polarity, modulation and composite probes

Two choices in Section II reflect a conveniences for deriving the observations in this work, and an additional step must be taken to expand the results out to hardware-relevant results. Thresholds were taken symmetric, $C_{+1} = C_{-1} = C$, and every probe considered is monotonically rising, so the analysis is carried entirely by ON events. Real devices bias the two polarities independently and the asymmetry is routinely tens of percent, so a probe with both rising and falling segments makes polarity an additional axis of $\boldsymbol{S}_i$ at no measurement cost, and the inverses must be derived per polarity. We leave that to future work, and note only that the derivation follows the same route as Section III rather than needing a new one.

The natural continuation is the periodic probe. Modulated illumination is already used across the field: for bias and event-rate control [26], for radiance recovery from event frequency [27], for frequency-tagged structured light [28], and for contrast-threshold calibration by a linear intensity ramp [29]. The contribution available there is not the stimulus but the correspondence: which statistic of the modulated stream carries the spatial transfer function, what its inverse is, and which sensor parameters that inverse requires. A companion observation on the temporal axis, that the frequancy components of an event stream attenuates coefficients below the peak signal frequency and injects non-physical harmonics [30], is the direct analogue of equation (18) and indicates what such an analysis has to control for.

Composite probes follow from the same treatment. A staircase (combining the step and ramp probes), a modulated square wave and a chirp trade dynamic range, frequency coverage, and correlated statistics against each other, and each admits the same chain of equation (16) once its statistic is nominated.

### E. Pixel models, simulators and hardware

The pixel model of Section II retains a single front-end pole. A second-order front end will directly influence the early intervals, since the transient is where the second pole lives. The linear-ramp derivation of Section III.C assumes that the single pole acts as a pure delay which cancels between consecutive events; a second pole breaks that cancellation, and the first-interval inverse acquires a term that does not divide out under equation (17). The question is whether the first-interval statistic remains the strongest of the three under a two-pole front end, or whether the rate statistic overtakes it.

Characterised-parameter simulators [18], [19] are not the route to re-deriving the inverses under a more complete model, which is unlikely to be tractable in closed form. Their use is to measure how far the closed-form inverse departs from the simulated truth as each non-ideality is switched on, which yields a validity envelope for each probe and statistic pair. The closed-form derived statistics should be used as first-order parameters for informed design (for instance), while elaborate models provide an envelope, rather than a new set of formulas, for these to work within under more realistic scenarios. This approach can be imperative for parametrising universal EVS pixel models [21], using agreed upon coefficients (analytical in first order, and experimental second order values). Hardware in the loop is where this coefficient-based modelling is trully benchmarked. Testing where any closed-form or simulated inverses holds is obviously needed. It is important to ensure in any such future work that readout-level effects (outside every pixel model) such as bus arbitration, timestamp quantisation and event-rate-dependent latency, are properly separated from the internal elements.

## VI. Conclusions

An event-measured PSF is not the optical one. It is a pointwise nonlinear warp of the optical weight, set jointly by the probe, the pixel model and the statistic read from the stream, and equation (16) states the order of operation that must be taken before any inference on the events can be made physically-relevant. Working that chain through for an idealised EVS sensor, three probes shows the inverses are not equally demanding: the step returns the relative amplitude and needs the full set of pixel parameters to undo an exponential, while the ramps return a time scale and need only a single additive offset, with the exponential ramp requiring no sensor parameter at all once its stationary interval and its phase are read separately. The practical aspect that follows this argument is that any event transfer function that acts to reduce the output dimension in relation to the event stream through a specific statistic – generally call a time-surface – must match excitation to inferred information.

## Acknowledgements

**Use of AI tools.** Claude Opus 5 (Anthropic) was used as an interactive collaborator. Its contributions: literature retrieval and cross-checking against primary sources; independent derivation and verification of the closed-form expressions; review and debugging of the simulation code; structured critique of the manuscript's argument; and drafting of manuscript prose from structure and content specified by the author. All AI-based derivations were checked by the author against the stated pixel model, all claims were verified numerically or traced to a cited source, and all scientific judgements, scope decisions and interpretations are the author's. The author takes full responsibility for the content. The simulation code for all numerical result is available online[1].

[1] `https://github.com/nimrodkruger/EVS_dynamic_transfer_functions.git`

# Supplementary Information - full derivation of the probe inverse functions

## A. The step probe

The step probe is $d(t) = Au(t - t_0)$, giving the per-pixel log-irradiance range $\Delta L_i = \log\left(1 + w_i A / I_{\text{bg}}\right)$. Time is measured from onset, $t_0 = 0$, throughout this section, matching the main text's convention for equation (23)-(26). Under the single-pole model in equation (4), the step response is

$$V(t) = \Delta L_i \left(1 - e^{\{-t/\tau_{\text{pr}}\}}\right), \quad t \geq 0. \tag{S1}$$

This is the standard first-order step response of $\tau_{\text{pr}} \dot{V} + V = \Delta L_i$ with $V(0) = 0$.

### A.1 Event count $N_i$

We write $x_k = e^{\{-t_k/\tau_{\text{pr}}\}}$ and $r = e^{\{-\rho/\tau_{\text{pr}}\}}$. No refractory window precedes the first event, so it fires the instant $V$ reaches $C$:

$$V(t_1) = C \Longrightarrow \Delta L_i (1 - x_1) = C \Longrightarrow x_1 = 1 - C/\Delta L_i, \tag{S2}$$

matching the initial condition quoted for equation (23). For $k \geq 1$, the reference resets to $V(t_k + \rho)$, and the $(k+1)$-th event fires when $V$ has climbed a further $C$ above it:

$$V\left(t_{\{k+1\}}\right) - V(t_k + \rho) = C. \tag{S3}$$

Substitute the step response into both terms. Since $e^{\{-(t_k+\rho)/\tau_{\text{pr}}\}} = r x_k$,

$$\Delta L_i \left(1 - x_{\{k+1\}}\right) - \Delta L_i (1 - r x_k) = C \Longrightarrow x_{\{k+1\}} = r x_k - C/\Delta L_i, \tag{S4}$$

which is equation (23), with $x_1 = 1 - C/\Delta L_i$ as above.

This is a linear (affine) recursion. We write $\beta \equiv C/\Delta L_i$, and equation (S4) becomes: $x_{\{k+1\}} = r x_k - \beta$, and it fixed point is

$$x^* = -\beta/(1 - r), \tag{S5}$$

found by setting $x^* = r x^* - \beta$. The deviation $\varphi_k \equiv x_k - x^*$ obeys the homogeneous recursion $\varphi_{\{k+1\}} = r \varphi_k$, so

$$x_k = x^* + (x_1 - x^*) r^{\{k-1\}}. \tag{S6}$$

The burst ends at the count $N_i$ where $x_{\{N_i\}}$ reaches zero, and the source-follower has settled and no further crossing occurs. Setting $x_N = 0$ and solving for $N$ (treated as continuous and taking the limit of the recursion) gives, after substituting $x_1 = 1 - \beta$,

$$r^{\{N-1\}} = \beta / [(1 - r) + \beta r] \Longrightarrow N = 1 + \tau_{\text{pr}}/\rho \cdot \log(r + \Delta L_i (1 - r)/C). \tag{S7}$$

This is exact for the recursion above. In the regime relevant to real sensors, many refractory periods elapse before the burst ends, $\rho \ll \tau_{\text{pr}}$, so $N_i \gg 1$ and $r = e^{\{-\rho/\tau_{\text{pr}}\}} \approx 1$ — the leading "+1" is negligible against $N_i$ itself, and $r$ may be replaced by 1 inside the logarithm (the exact refractory factor $(1 - r)$ is kept, not linearised). Equation S7 is then reduces to Equation (24) from the main text,

$$N_i \approx \tau_{\text{pr}}/\rho \cdot \log\left(1 + \Delta L_i \left(1 - e^{\{-\rho/\tau_{\text{pr}}\}}\right)/C\right). \tag{S8}$$

Both forms agree closely once $\rho/\tau_{\text{pr}} \ll 1$.

### A.2 Onset latency $\ell_i$

This derivation needs only the first-event condition from A.1, with no recursion. With $\ell_i \equiv t_1$,

$$V(\ell_i) = C \Longrightarrow \Delta L_i \left(1 - e^{\{-\ell_i/\tau_{\text{pr}}\}}\right) = C \Longrightarrow \Delta \hat{L}_i = C / \left(1 - e^{\{-\ell_i/\tau_{\text{pr}}\}}\right), \tag{S9}$$

exactly the second case of Equation (26).

### A.3 First interval $\Delta t_{\{i,1\}}$

We apply the general recursion once, at $k = 1$: $x_2 = r x_1 - \beta$. Writing $x_2 = x_1 e^{\{-\Delta t_{\{i,1\}}/\tau_{\text{pr}}\}}$ (since $t_2 = t_1 + \Delta t_{\{i,1\}}$) and substituting $x_1 = 1 - \beta$,

$$(1 - \beta) e^{\{-\Delta t_{\{i,1\}}/\tau_{\text{pr}}\}} = r(1 - \beta) - \beta. \tag{S10}$$

Let $q \equiv r - e^{\{-\Delta t_{\{i,1\}}/\tau_{\text{pr}}\}}$ (non-negative, since $\Delta t_{\{i,1\}} \geq \rho$). Rearranging the line above gives $\beta = (1 - \beta) q$, so

$$\beta = q/(1 + q) \Longrightarrow \Delta L_i = C(1 + 1/q). \tag{S11}$$

Substituting $q$ back gives Equation (26)'s third case exactly,

$$\Delta \hat{L}_i = C\left(1 + 1/\left(e^{\{-\rho/\tau_{\text{pr}}\}} - e^{\{-\Delta t_{\{i,1\}}/\tau_{\text{pr}}\}}\right)\right). \tag{S12}$$

## B. The linear ramp

The ramp is $d(t) = \alpha(t - t_0)$ (27). Writing $s = t - t_0$ and collecting the per-pixel terms into a characteristic time,

$$T_i \equiv I_{\text{bg}}/(w_i \alpha), \tag{S13}$$

the irradiance is $I_{i(s)} = I_{\text{bg}} + w_i \alpha s = I_{\text{bg}}(1 + s/T_i)$, so, as in Equation (29),

$$L_{i(s)} = \log I_{\text{bg}} + \log(1 + s/T_i), \tag{S14}$$

a function of $s/T_i$ alone: the weight sets a **time scale**, not an amplitude, unlike the step. Every statistic below estimates $T_i$, then recovers $w_i = I_{\text{bg}}/(\alpha T_i)$.

### B.1 Characteristic time and the first-interval recursion

Applying the crossing condition in Equation (5) to Equation (29), with the reference taken at the end of the refractory window, $V_{i\left(s_{\{k+1\}}\right)} - V_{i(s_k+\rho)} = C$:

$$\log\left(1 + s_{\{k+1\}}/T_i\right) - \log(1 + (s_k + \rho)/T_i) = C \Longrightarrow \left(T_i + s_{\{k+1\}}\right)/(T_i + s_k + \rho) = e^C, \tag{S15}$$

recovering the main text's Equation (30) directly. Define $u_k = T_i + s_k$, and Equation (30) becomes the linear recursion

$$u_{\{k+1\}} = e^C (u_k + \rho). \tag{S16}$$

Before onset the irradiance is constant, so the reference has settled at $\log I_{\text{bg}}$ and the first event carries no preceding refractory window: $u_1 = T_i e^C$ (equivalently $s_1 = T_i\left(e^C - 1\right) > 0$).

Solve as in A.1: the fixed point is $u^* = e^C \rho / \left(1 - e^C\right) = -e^C \rho / \left(e^C - 1\right)$, and $u_k = u^* + (u_1 - u^*) e^{\{C(k-1)\}}$.

Writing $\tilde{T}_i \equiv (u_1 - u^*) e^{\{-C\}}$,

$$\tilde{T}_i = T_i e^C \cdot e^{\{-C\}} - u^* e^{\{-C\}} = T_i + \rho / \left(e^C - 1\right), \tag{S17}$$

using $u^* e^{\{-C\}} = -\rho/\left(e^C - 1\right)$, and

$$u_k = \tilde{T}_i e^{\{kC\}} - \rho e^C/\left(e^C - 1\right). \tag{S18}$$

The interval $\Delta t_{\{i,k\}} = u_{\{k+1\}} - u_k = \tilde{T}_i e^{\{(k+1)C\}} - \tilde{T}_i e^{\{kC\}} = \tilde{T}_i e^{\{kC\}}\left(e^C - 1\right)$, as Equation (31).
For the first interval, $\Delta t_{\{i,1\}} = u_2 - u_1 = e^{C(u_1+\rho)} - u_1 = u_1\left(e^C - 1\right) + \rho e^C$. With $u_1 = T_i e^C$,

$$\Delta t_{\{i,1\}} = e^C\left[T_{i(e^C-1)} + \rho\right]. \tag{S19}$$

Solving for $T_i$ and then $w_i = I_{\text{bg}}/(\alpha T_i)$ gives Equation (32),

$$\hat{w}_i = \left(I_{\text{bg}} e^C\left(e^C - 1\right)\right)/\left(\alpha\left(\Delta t_{\{i,1\}} - \rho e^C\right)\right). \tag{S20}$$

**B.2 Onset latency $\ell_i$**

As noted in the main text, once $\tau_{\text{pr}} \ll T_i + s$ the single-pole filter acts as a pure delay, $V(s) \approx L\left(s - \tau_{\text{pr}}\right)$: the front-end pole common to every crossing (array uniformity assumed) shifts the **observed** onset time by $\tau_{\text{pr}}$ without otherwise altering Equation (29). The undelayed crossing of $L_i$ itself, $L_{i(s)} - \log I_{\text{bg}} = C$, gives from Equation (29)

$$\log(1 + s/T_i) = C \Longrightarrow s = T_i\left(e^C - 1\right) \tag{S21}$$

(consistent with $s_1$ above). The **measured** latency is this plus the front-end delay, $\ell_i = T_{i(e^C-1)} + \tau_{\text{pr}}$, so

$$T_i = \left(\ell_i - \tau_{\text{pr}}\right)/\left(e^C - 1\right) \Longrightarrow \hat{w}_i = I_{\text{bg}}/\alpha \cdot \left(e^C - 1\right)/\left(\ell_i - \tau_{\text{pr}}\right), \tag{S22}$$

and exactly Equations (33)‘s latency case.

**B.3 Rate $R_i$**

$R_i$ is the mean of $1/\Delta t_{\{i,k\}}$ over the $N_i - 1$ intervals following onset,

$$R_i = (N_i - 1)^{\{-1\}} \sum_{\{k=1\}}^{\{N_i-1\}} 1/\Delta t_{\{i,k\}} \tag{S23}$$

.

From B.1, $1/\Delta t_{\{i,k\}} = e^{\{-kC\}}/\left(\tilde{T}_i\left(e^C - 1\right)\right)$, so with $M = N_i - 1$,

$$R_i = \frac{1}{\tilde{T}_i(e^C - 1)M} \sum_{\{k=1\}}^{\{M\}} e^{\{-kC\}}. \tag{S24}$$

The sum is a finite geometric series. Multiplying numerator and denominator of the standard sum by $e^C$,

$$\sum_{\{k=1\}}^{\{M\}} e^{\{-kC\}} = e^{\{-C\}} \frac{1 - e^{\{-MC\}}}{1 - e^{\{-C\}}} = \frac{1 - e^{\{-MC\}}}{e^C - 1}. \tag{S25}$$

This relation is purely an algebraic step, and is the ordinary closed form for $\sum_{\{k=1\}}^{M} r^k$ with a common ratio $r = e^{\{-C\}}$, rewritten so the denominator matches the $\left(e^C - 1\right)$ already carried by every $\Delta t_{\{i,k\}}$. Substituting back,

$$R_i = \frac{1 - e^{\{-(N_i - 1)C\}}}{\tilde{T}_i (e^C - 1)^2 (N_i - 1)} = \Gamma(N_i)/\tilde{T}_i, \tag{S26}$$

where

$$\Gamma(N) \equiv \frac{1 - e^{\{-(N-1)C\}}}{(N-1)(e^C - 1)^2} \tag{S27}$$

is exactly the correction factor quoted in Equation (33) - an average of the geometric series of inverse intervals. Recovering $T_i = \tilde{T}_i - \rho/(e^C - 1)$ and then $w_i = I_{\text{bg}}/(\alpha T_i)$ gives Equation (33)'s rate case,

$$\hat{w}_i = I_{\text{bg}}/\alpha \cdot \left[\Gamma(N_i)/R_i - \rho/(e^C - 1)\right]^{\{-1\}}. \tag{S28}$$

Unlike A.1′s count, no "many events" approximation is needed anywhere in B.1-B.3, therefore the ramp's recursion is exactly solvable at every step because its multiplier $e^C$ is constant, whereas the step's multiplier $r$ competes with an additive term of comparable size only for a **finite** burst.

## C. The exponential ramp

The exponential ramp is $d(t) = A\left(e^{\left\{\frac{t-t_0}{\tau_\alpha}\right\}} - 1\right)$ from Equation (34), giving, with $s = t - t_0$, the main text's Equation (35),

$$L_{i(s)} = \log\left[(I_{\text{bg}} - w_i A) + w_i A e^{\left\{\frac{s}{\tau_\alpha}\right\}}\right]. \tag{S29}$$

### C.1 Onset latency $\ell_i$

No refractory window precedes the first event, so it fires the instant $L_i$ has risen by $C$ from $\log I_{\text{bg}}$:

$$L_{i(\ell_i)} - \log I_{\text{bg}} = C \Longrightarrow \log\left[\frac{(I_{\text{bg}} - w_i A) + w_i A e^{\left\{\frac{\ell_i}{\tau_\alpha}\right\}}}{I_{\text{bg}}}\right] = C. \tag{S30}$$

Exponentiate and clear $I_{\text{bg}}$:

$$(I_{\text{bg}} - w_i A) + w_i A e^{\left\{\frac{\ell_i}{\tau_\alpha}\right\}} = I_{\text{bg}} e^C \Longrightarrow w_i A\left(e^{\left\{\frac{\ell_i}{\tau_\alpha}\right\}} - 1\right) = I_{\text{bg}}(e^C - 1). \tag{S31}$$

Solving for $\ell_i$ gives Equation (38) exactly, with no asymptotic approximation:

$$\ell_i = \tau_\alpha \log\left(1 + I_{\text{bg}} \frac{e^C - 1}{w_i A}\right). \tag{S32}$$

Inverting the boxed relation above directly for $w_i$ gives Equation (39),

$$\hat{w}_i = I_{\text{bg}} \frac{e^C - 1}{A\left(e^{\left\{\frac{\ell_i}{\tau_\alpha}\right\}} - 1\right)}, \tag{S33}$$

an exact algebraic inversion of Equation (35) at one point.